# Leveraging LLMs for Structured Data Extraction from Unstructured Patient Records

Mitchell A. Klusty[1], BS[1,†], Elizabeth C. Solie[1,†], Caroline N. Leach[1], BS[1], W. Vaiden Logan, BS[1], Lynnet E. Richey, BS[2], John C. Gensel, PhD[2], David P. Szczykutowicz, BA[2], Bryan C. McLellan, BS[2], Emily B. Collier, MS[1], Samuel E. Armstrong, MS[1], V. K. Cody Bumgardner, PhD[1]

[1]Center For Applied Artificial Intelligence, University of Kentucky, Lexington, KY
[2]Spinal Cord and Brain Injury Research Center and Department of Physiology, University of Kentucky College of Medicine, Lexington, KY
†These authors contributed equally and share first authorship.

**Abstract**

*Manual chart review remains an extremely time-consuming and resource-intensive component of clinical research, requiring experts to extract often complex information from unstructured electronic health record (EHR) narratives. We present a secure, modular framework for automated structured feature extraction from clinical notes leveraging locally deployed large language models (LLMs) on institutionally approved, Health Insurance Portability and Accountability Act (HIPPA)-compliant compute infrastructure. This system integrates retrieval augmented generation (RAG) and structured response methods of LLMs into a widely deployable and scalable container to provide feature extraction for diverse clinical domains. In evaluation, the framework achieved high accuracy across multiple medical characteristics present in large bodies of patient notes when compared against an expert-annotated dataset and identified several annotation errors missed in manual review. This framework demonstrates the potential of LLM systems to reduce the burden of manual chart review through automated extraction and increase consistency in data capture, accelerating clinical research.*

**Introduction**

Electronic Health Records contain a wealth of unstructured textual data, including physician notes, discharge summaries, and consultation reports. These narrative documents are rich in clinical detail, capturing nuances of patient histories, diagnostic reasoning, treatment responses, and care planning that are often absent from structured fields. However, the unstructured nature of these records presents significant challenges for areas where they are most useful such as clinical decision support, cohort identification for research studies, quality improvement initiatives, and population health analytics. These tasks often require structured, machine-readable information, necessitating reliable extraction of relevant features from the unstructured EHRs. Historically, this extraction has been performed either by manual annotation or rule-based systems, both of which are labor-intensive, error-prone, and difficult to scale when applied to a high volume of patients.

Recent advances in LLMs have opened new opportunities for automating clinical information extraction with unprecedented accuracy and flexibility. LLMs are deep neural networks trained on massive corpora of human language and can perform a wide range of natural language processing (NLP) tasks with little to no task-specific training. In the clinical domain, they offer the ability to generalize across diverse documentation styles, specialty-specific vocabularies, and note structures, greatly reducing the need for handcrafted rules or large labeled datasets. This has the potential to significantly accelerate the traditionally slow and manual process of transforming unstructured clinical narratives into structured data, making information more accessible, actionable, and useful for both patient care and secondary uses of health data.

To harness this potential, we present a scalable, clinician-friendly tool that leverages open-source LLMs to automate the extraction of structured features from EHR notes, with rigorous protections to safeguard sensitive patient data, with rigorous protections to safeguard sensitive patient data. The system processes heterogeneous patient data originating from institutional EHR platforms like Epic[1], accessed through the institution's REDCap[2,3] environment, which provides secure, audit-controlled data management. Our system uses various open-source models such as *Qwen/QwQ-32B*[4] to perform inference within an on-premises, HIPAA-compliant[5] infrastructure, ensuring data privacy and regulatory adherence. Output structure is governed by customizable and OpenAI-formatted tools[6] that describe the desired schema, enabling consistent formatting across various clinical applications. Where available, we employ these tool calling features to improve the precision and flexibility of field extraction.

Tool calling was originally designed as a mechanism for enabling LLMs to execute external functions, identifying relevant pieces of input text and supplying it as arguments for the function calls, similar to filling in blanks on a form to instruct the program. Although this capability was created for software automation, it also provides a powerful and intuitive way to perform structured feature extraction from clinical text. One of the central challenges in clinical text analysis is that patient notes are written in natural language. Computers work best when information is provided in a standard, structured format. Tool calling bridges the gap between rich, narrative, unstructured text and machine-readable, structured, database-ready information.

In our system, the LLM is prompted with a set of clinical notes and instructed to provide a specific set of features (e.g. diagnoses, dates, symptoms, etc.) that it can identify from those notes as a machine-readable JSON object, mirroring the schema for a function call in software development. This utilizes the natural language processing capabilities of LLMs to enforce structural consistency across outputs, allowing the model to populate structured clinical variables such as diagnosis, dates, and supporting evidence, much like a physician reading the notes and filling in the fields in a form. With the data in this structured format, it can be reliably stored in a database, used for statistical analysis, or integrated into other components of clinical decision support processes.

Importantly, the system is designed to be accessible to clinicians without technical backgrounds. Through a streamlined graphical interface, users can define the fields they wish to extract using simple, intuitive forms. This removes the need for programming or familiarity with model prompting techniques. Once configured, the system securely processes patient notes and produces structured outputs that can be directly integrated into clinical workflows, research pipelines, or data warehousing platforms.

This work demonstrates how state-of-the-art open-source LLMs can be operationalized in a practical and compliant manner to unlock valuable insights from unstructured clinical documentation. By bridging the gap between complex AI models and real-world healthcare needs, our tool empowers clinicians and researchers to make better use of their data with minimal technical overhead.

**Methods**

The framework we developed was designed to be modular and scalable to support automated extraction of structured features from unstructured clinical notes using LLMs. Deployment of this framework is intended to operate within HIPAA-compliant environments, provide secure data management, support reproducibility of results, and be adaptable to diverse institutional computing infrastructure.

The architecture is composed of four primary components: (1) integration with clinical data sources via REDCap, (2) secure data transfer and management processes, (3) containerized LLM embedding of reports and feature searching and extraction, and (4) a web interface for definition of extraction parameters and initialization of extraction jobs. This section describes the technical implementation for these components.

*1. Clinical Data Sources and REDCap Integration*

Patient data was obtained from multiple institutional EHR systems: EPIC, which is used for contemporary data, Sunrise Clinical Manager (SCM)[7] and Allscripts Electronic Health Records (AEHR)[8] for legacy notes predating EPIC integration. The data from these platforms contains both structured and unstructured clinical information. These datasets contain medical record numbers (MRNs), encounter timestamps, and associated clinical text fields, including encounter summaries, imaging reports, laboratory results, and narrative clinical notes. All data was exported from these sources using institutionally approved and standardized procedures into the university's instance of REDCap, which serves as the centralized data repository for our system. REDCap was selected for its robust data access controls, granular audit capabilities, and Application Programming Interface (API)-based programmatic integration features. This facilitates reproducible and traceable data workflows in compliance with institutional privacy governance, critical components when processing Protected Health Information (PHI).

Each dataset retrieved from REDCap by our framework is exported as a structured tabular data file (CSV or XLSX), with each record linked to a patient MRN. These MRNs serve as unique, stable identifiers across the various heterogeneous data sources, allowing all data to be aligned and aggregated for individual patients.

*2. Secure Data Transfer and Local Mounting to Compute Container*

Secure transportation and storage of data is essential when PHI is present. To guarantee this, all data transfers and processing occur within HIPAA-compliant environments. Access to REDCap data is managed through the REDCap API. Rather than providing direct database access or requiring user interaction with the web frontend, the API acts as a controlled gateway whereby authorized users are issued unique API tokens that provide programmatic access to specify exactly what data they can retrieve or modify. Each request through the API is automatically logged to support full auditability of all transactions.

At the start of the containerized feature extraction process (*Methods, Section 3*), the note files are downloaded to the compute node through the REDCap API, and once extraction is complete, the results are exported back to REDCap through the API. When starting an extraction job, the user provides their API token to the system. It is securely stored as an environment variable and used for all data transfers to and from REDCap. If utilizing the web frontend (*Methods, Section 4*), for extraction job management, the token is supplied once by the user at job initialization, encrypted, and forwarded directly to the container. Upon completion of the extraction, the token is deleted to prevent possible credential leakage, which would allow unauthorized access to protected data by impersonating an authorized user. This methodology maintains tight security over the token.

All transfers occur over encrypted HTTPS connections, protecting the sensitive information in transmission. Transferred files are stored on local volumes on the compute node and mounted directly to the extraction container. This design permits a secure, auditable, and automated transfer protocol that maintains compliance with institutional data governance and privacy requirements.

## 3. Containerized LLM Batch Processing Pipeline

Extraction jobs are performed as batch inference within Docker containers[9] intended for computational clusters that use the Slurm[10] workload manager. Docker-based containers ensure reproducibility and environment isolation with package dependencies versioned through internal registries. Each job launches a containerized inference environment using the vLLM[11] runtime, serving a locally hosted, configurable LLM. This means the model operates entirely on institutional servers rather than on an external cloud service, allowing all PHI to remain within the organization's existing security perimeter while giving administrators control over how the model is tuned, accessed, and audited. The inference framework is modular and follows standard OpenAI API formatting, allowing seamless configuration of various models without requiring modifications to the processing pipeline. The framework supports many open-source models hosted on Hugging Face[12], including the Llama[13] and Qwen model families.

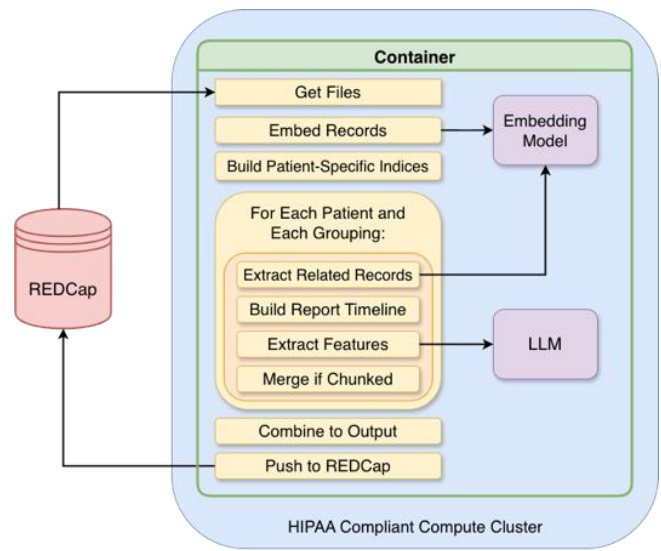

*Figure 1 The design of the batch inference container*

A major challenge presented in this process is that individual patients often have an extensive volume of clinical notes, which can exceed the context length of current large language models. The context length of a model defines the maximum number of tokens (roughly equivalent to words or sub-word units) it can process at once in both its input and output. Most contemporary models support up to approximately 128,000 token context lengths. However, despite this nominal capacity, there is often a marked decline in model comprehension and factual consistency as the number of input tokens increases, as noted by Liu et al.[14] To accommodate these limitations, we implemented a preprocessing stage that removes extraneous content known to provide no clinical value, such as long XML metadata sections or system generated headers that do not contain patient notes. After cleaning, the data can often still be several million tokens, which is still too large for efficient, reliable extraction, particularly on more limited compute infrastructure. To address this, we split the extraction process into two steps: (1) using RAG[15] methods to limit the search to only relevant reports and (2) chunked analysis of filtered reports.

*3.1 Vector Database Searching to Limit Report Context*: One popular method of RAG is using text embeddings to create a vector database, which can be searched to identify documents that have similar meaning. This entails using a language transformer model to convert each report into a semantic vector representation that enables similarity searching. Reports for each individual MRN in the datasets are embedded with a configurable LLM-based embedding model, such as *sentence-transformers/all-mpnet-base-v2*[16], and the generated embeddings are L2-normalized to support efficient comparison using cosine similarity metrics. The normalized embeddings are stored in individual FAISS index files[17] and Parquet metadata files[18], serving as vector databases which allow semantic retrieval of reports that are most relevant to the target extraction features. Reports that exceed the context length of the embedding model (often around 1024 tokens) are split into overlapping chunks to ensure all data is embedded without losing context at the edge of the chunks.

The user splits the target features into logical groupings that should be analyzed together. For instance, if the user wishes to identify if a patient has had a stroke and the date of the stroke, those features would be logically grouped, but if they also want to identify if the patient is a smoker, that would be a second group. An LLM generates a strong set of search terms that relate to each individual feature grouping, embeds those features, normalizes the generated embeddings using L2-normalization, which simply rescales the vectors so they have a magnitude of 1. This makes cosine similarity comparisons dependent only on the direction of the embeddings, which maintains their semantic meaning. The system then performs a cosine similarity search to identify all related documents with a higher calculated similarity than a configurable threshold. Because the embeddings are normalized, the cosine similarity scores follow a distribution such that all scores fall in the range [-1, 1] where a score closer to 1 indicates a stronger semantic similarity, while a score closer to -1 indicates very little semantic relation. All reports with a calculated score greater than or equal to the configured threshold are selected for analysis for that feature group.

*3.2 Feature Extraction from Filtered Reports*: The remaining, filtered reports are provided as context to the LLM to perform feature extraction. As described above, the tabular data encompasses multiple structural formats originating from different EHR systems. To enable effective analysis by an LLM, this data must be arranged into a unified representation that provides longitudinal context across each patient's history. For this purpose, when providing the notes to the LLM for feature extraction, a single, consolidated report is generated for each of the selected reports. These consolidated reports aggregate and chronologically order the patient's encounters, imaging and lab results, and notes, constructing a comprehensive timeline of clinical events tracked in the data.

For a single analysis, the model must process the system prompt (instructions provided to explain the setting and desired output), the tool definition, the entire text of the collected reports, and the structured output. The total tokens must not exceed the context length of the chosen model. If the consolidated report is still too long, it must be split in a controlled manner to avoid exceeding that limit. In such cases, the report is divided into smaller, overlapping, sequential chunks that fit within the model's context window. Each chunk is processed independently by the LLM, and the resulting output features are subsequently merged through a rule-based reconciliation methodology to produce a unified analysis across the patient's full timeline.

As the extraction of each feature grouping from individual patient records is complete, the results are saved to a CSV file, creating a checkpointing system that allows restart on error or interruption without requiring reprocessing of previously analyzed data. Upon analysis of each unique record, the data is exported back to REDCap as a CSV.

The system is designed to be infrastructure-agnostic and scalable to available compute resources, with a minimum requirement that the compute node has enough GPU memory to support the configured LLM. As analysis jobs run inside a Docker container, all dependencies, the user-provided configuration, and the runtime environment and resources are encapsulated and isolated. The container automatically initializes a vLLM runtime with the configured model and GPU allocations, supporting efficient inference with parallelizable calls that scale with available resources.

The framework is compatible with both direct Docker execution and Slurm-based batch job scheduling. When deployed on a compute cluster, users can submit feature extraction jobs as Slurm tasks. Slurm is an open-source workload manager widely used in high-performance computing (HPC) environments to allocate resources, queue jobs, and coordinate distributed execution. This allows resource-aware scheduling and queue management, as well as parallel execution over multiple nodes, supporting execution on diverse, HIPAA-compliant compute infrastructures, from single-node GPU servers to multi-node clusters.

The LLM and embedding models are open-weight, locally hosted, available through Hugging Face, and configured in an OpenAI API-compatible format, allowing easy interchangeability between models. This modularity allows for strong extensibility as new models and supporting frameworks are developed without requiring changes to the core processing logic.

*4. Web Frontend*

To make the system accessible to clinicians and researchers, who often lack technical expertise with utilizing LLM systems, we developed a dedicated web-based frontend that offers an intuitive interface for defining feature extraction schemas and configuring and initiating extraction processes using those schemas on the distributed compute infrastructure.

*4.1 Extraction Tool Definition*: Rather than requiring users to manually write JSON schemas or complex prompts, the interface wraps the OpenAI tool-calling specification through a guided, form-based configuration workflow. This allows users to translate clinical knowledge and decision-making processes into valid, OpenAI-formatted tools for structured extraction logic through high-level, front-end interactions.

Users define the clinical features they wish to extract from patient notes by creating a *Tool* composed of modular *Fields*. For each *Field*, users specify the name, description, data type (e.g., string, number, integer, Boolean, array, or object), and whether it is required. Required *Fields* indicate that the extraction process must always attempt to retrieve the value, and the model considers the extraction incomplete if it is not found. Additional constraints, such as enumerated options for a field, regex patterns for strings, minimum/maximum bounds for numeric fields, and default values, allow users to encode clinical feature guardrails without requiring additional code.

This configuration design supports both simple and more hierarchical data structures. For instance, a user may define a *Tool* containing simple *Fields* such as "Diagnosis" (string), "Age" (integer), and "Smoker Status" (Boolean). More complex structures can be modeled by nesting *Fields* within objects or arrays. A "Blood Pressure" *Field* may be represented as an object containing "Systolic" and "Diastolic" subfields, and a medication list can be defined as an array of objects with subfields for "Name" and "Dosage". This allows extraction of basic and more complex features from the patient notes, mirroring the variety of data and relationships that are common in clinical documentation.

*Figure 2 User interface form for creating a* Tool

The *Tool* structure is compiled into a fully defined OpenAI-compatible JSON schema. This schema serves as the formal tool description utilized by the backend LLM, promoting standardized, cross-model feature extraction. By utilizing this schema, the system reliably receives a reproducible, type-safe, user-constrained, structured output while minimizing the technical barriers for such LLM interaction. In practice, this frontend functions as a clinically oriented wrapper around the OpenAI tool-calling framework, allowing domain experts to define structured extraction tasks through intuitive interactions rather than technical specification.

*4.2 Job Initiating*: As previously discussed, because the patient data for this system are generally subject to HIPAA protection, the feature extraction pipeline must be performed within institutionally secured compute environments. To enable accessibility to the backend system without exposing PHI, the front-end is designed to exist externally from the HIPAA environment, only communicating metadata and configuration parameters to the backend. This allows the front-end to act independently of patient data, functioning as a job orchestration interface, while the backend container performs all operations that interact directly with PHI.

Users initiate an extraction job by specifying the *Tool* to be used for extraction, selecting the embedding and LLM models for inference, and providing delegated access to the patient note source files in REDCap. The extraction container is launched with these parameters set as environment variables. As described in *Methods, Section 2*, the user supplies the paths of the files in REDCap targeted for feature extraction, and an API token with access to those files is used to securely transport files to the compute node. The extraction process is then executed: the LLM pipeline processes retrieve the specified feature set, and that feature set is exported to REDCap. Upon completion, the container is torn down, deleting the API token from the compute environment to prevent credential reuse, and the web frontend is notified that the extraction is completed. The initial provisioning of configuration for the container and this notification of status are the only communication between the frontend and backend, providing a strict separation in the layers that prevent the frontend from ever interacting with PHI.

*4.3 User Access Management*: Though the front-end system is intentionally designed without direct access to patient data, strong authentication and authorization controls are still essential for user-defined *Tools* and extraction jobs. Authentication is handled through CILogon[19], an OAuth 2.0[20] identity provider that federates login credentials across academic institutions. CILogon connects directly to institutional single sign-on (SSO) systems, allowing our frontend to leverage established identity verification processes, multi-factor authentication (MFA) protocols and institutionally governed account management.

Authorization to specific resources within the platform is managed by a locally maintained role-based access management (RBAC) system. *Tools* and extraction jobs are provided three-tiers of roles with ascending permissions, **read**, **write**, and **manage,** which define the actions a user may take on these resources.

- **Read** access on a *Tool* allows the user to view the *Tool*'s structure and initiate an extraction job with it. On extraction jobs, **read** access permits viewing the status.
- **Write** access on a *Tool* includes **read** privileges and allows editing the *Tool*'s fields and configuration.
- **Manage** access grants full control over it, including delegation of permissions for the *Tool* or extraction job.

When a *Tool* is created or a job launched, the author is given **manage** permissions and may assign or revoke access to other users. This supports collaborative development across research teams within the system while providing an RBAC-based model to ensure proper authorization to user-generated resources.

**Results**

To evaluate the performance of this system, we used a curated dataset obtained from the University of Kentucky's EHR repositories. A reference, "gold-standard" set of patient-level feature annotations was independently created by domain experts from the Spinal Cord and Brain Injury Research Center (SCoBIRC). These experts manually reviewed and annotated target features from the patient notes, making a benchmark set of 100 patients that was compared to results extracted by the system. For this experiment, the container was provided one NVIDIA A100 GPU for inference to gauge the system's performance on comparatively low compute resources. The embedding model used was *sentence-transformers/all-mpnet-base-v2*, chosen for its strong relationship capturing capabilities[16]. The selected LLM was *Qwen/QwQ-32B*, chosen for its competitive evaluation metrics and strong reasoning capabilities.[5] Also, with tool calling enabled for the vLLM deployment, *Qwen/QwQ-32B* at 16-bit quantization occupies approximately 71GB of the available 80 GB on the provided A100 GPU.

The assessment focused on five clinically relevant feature groupings important for study participant criteria filtering: Traumatic Spinal Cord Injury (SCI), Myocardial Infarction (MI), Stroke, Type 2 Diabetes Mellitus (T2DM), and Transient Ischemic Attack (TIA). These groups were each comprised of a set of structured elements that were the targets for extraction by the system. Table 1 summarizes those features.

| Feature Group | Extracted Elements | Definition |
| --- | --- | --- |
| Traumatic Spinal Cord Injury (SCI) | Occurrence | Whether or not the patient is noted as having sustained a traumatic SCI |
| | Date | Earliest documented date of SCI |
| Myocardial Infarction (MI) | Occurrence | Whether the patient has ever had a myocardial infarction. |
| | Date | Earliest documented MI date. |
| Stroke | Occurrence | Whether the patient has ever had a stroke. |
| | Date | Earliest documented stroke date. |
| Transient Ischemic Attack (TIA) | Occurrence | Whether the patient has ever had a TIA. |
| | Date | Earliest documented TIA date. |
| Type 2 Diabetes Mellitus (T2DM) | Occurrence | Whether the patient has a diagnosis of T2DM. |
| | Date | Earliest documented diagnosis date. |

*Table 1 Breakdown of features extracted from patient notes to assess quality of extraction system*

Using the extraction methodology detailed in *Methods, Section 3*, the system processed all patient notes associated with the gold-standard dataset. For each patient, the model extracted the targeted features as structured output which was then saved to a CSV file and exported securely back to REDCap. We computed the agreement metrics between the expert annotations and the model-extracted features by aligning value agreement of Boolean fields (i.e. if the expert and the model agreed an SCI occurred for a particular patient) and determining if the year from extracted dates was agreed upon. We calculated the precision, recall, and F1-score metrics for each of the metrics, which are presented in Table 2.

| Feature Group | Feature | Precision | Recall | F1-Score | Accuracy | TP | TN | FP | FN |
|---|---|---|---|---|---|---|---|---|---|
| Traumatic Spinal Chord Injury | Occurrence | 0.8413 | 0.8689 | 0.8548 | 0.82 | 53 | 29 | 10 | 8 |
| | Date | 0.5645 | 0.8140 | 0.6667 | 0.65 | 35 | 30 | 27 | 8 |
| Myocardial Infarction (MI) | Occurrence | 0.7000 | 0.7778 | 0.7368 | 0.95 | 7 | 88 | 3 | 2 |
| | Date | 0.200 | 1.000 | 0.3333 | 0.92 | 2 | 90 | 8 | 0 |
| Stroke | Occurrence | 0.2143 | 0.5000 | 0.3000 | 0.86 | 3 | 83 | 11 | 3 |
| | Date | 0.1538 | 0.6667 | 0.2500 | 0.88 | 2 | 86 | 11 | 1 |
| Transient Ischemic Attack (TIA) | Occurrence | 0.4000 | 1.0000 | 0.5714 | 0.97 | 2 | 95 | 3 | 0 |
| | Date | 0.0000 | 0.0000 | 0.0000 | 0.96 | 0 | 95 | 5 | 0 |
| Type 2 Diabetes Mellitus (T2DM) | Occurrence | 0.8333 | 0.7895 | 0.8108 | 0.93 | 15 | 78 | 3 | 4 |
| | Date | 0.0000 | 0.0000 | 0.0000 | 0.85 | 0 | 85 | 14 | 1 |

*Table 2 Results of the extraction agreement with expert-labeled "gold-standard" dataset.*
*TP=True Positive, TN=True Negative, FP=False Positive, FN = False Negative*

False positives, in this case, indicate the model found an affirmative result that the expert annotation labeled to the negative. For Booleans, this means the model extracted *true* for the field where the expert labeled *false* (e.g. the model indicated there was an SCI while the annotator found there was not according to the provided definition of SCI). For date fields, this means the model extracted a date, and the year matched the expert annotated year. False negatives indicate cases where no feature was extracted, but the expert annotation provided a label. For Booleans, this means the model extracted *false* where the manual label was *true*. For dates, the model did not find a date, but the expert label did.

The lowest performing extracted field is the Traumatic SCI date. Aside from this field, each of the extractions performed with >82% agreement with the manually annotated data. The primary sources of error were false positives, with a small number of false negatives across the feature groups. Qualitative assessment of the individual disagreements indicated that many differences stemmed from underdeveloped or incomplete instruction prompts or tool descriptions, especially in cases where additional clinical context (e.g. severity of SCI that should be included) was needed to guide the model's reasoning. When prompts were refined to provide stricter criteria to the extraction, precision and recall improved, showing the important role of prompting in extraction quality.

In interpreting these results, it is important to acknowledge the inherent ambiguity that exists in generating expert-labeled "gold-standard" datasets of this kind. Determining the earliest date of a condition is often not straightforward, as EHR narratives often reference historical events, prior hospitalizations, or findings documented second-hand in subsequent encounters. For example, notes may state that a traumatic injury occurred "about a decade ago" or when the patient "was a young child". Such descriptions led annotators to record an approximate date of diagnosis, while the program yielded false negatives. This prevarication in the source documents introduces unavoidable uncertainty to date-level annotations, which makes the ground truth for those fields less absolute than Boolean or more discrete variable fields. These ambiguities highlight the difficulty of manual review across vast sets of unstructured EHR data and suggest a disagreement between the model's output and the annotation do not necessarily indicate an error in the system.

Despite these challenges, the system showed strong performance in identifying true negatives across all feature groups, meaning it rarely ascribed conditions to patients where no strong evidence existed in the notes. The Traumatic SCI group provides a particularly illustrative example in comparison to the other feature groups, as there were a higher proportion of patients who suffered from this condition than the others. As a result, its metrics provide a more balanced view of positive and negative cases, making it more sensitive to recall and precision. The performance of the model on this feature group, while slightly lower than the more sparse conditions, demonstrates the model maintains reasonable accuracy, even when there are more distinguishing characteristics and nuance in arriving at a positive or negative finding.

During evaluation, several apparent errors in the system's output were found to be the result of omissions or inaccurate labels in the gold-standard annotations. For a small set of cases, the model correctly identified clinical events or diagnoses that had been missed by human annotators. These instances were subsequently reviewed by domain experts and confirmed as true positives. This validation shows not only the complexity of current manual chart review procedures, but also the potential of LLM-assisted review systems to both significantly increase speed in review completion and provide quality assurance for reviewed datasets.

**Discussion**

**The results of this evaluation demonstrate the ability of the proposed framework to reliably extract clinical features from EHR text using LLMs deployed locally in a HIPAA compliant environment, achieving high accuracy across evaluated feature groupings. The system's modular architecture, which utilizes RAG methods for report filtering and structured tool-based LLM extraction in a secure, containerized inference pipeline, supports consistent performance across diverse clinical feature sets and broad deployability for available institutional compute resources.** Most modern LLMs perform very well with simple feature extraction, and there is literature supporting the ability of more powerful LLMs to extract complex characteristics as well,[21] when provided full context for that feature[15]. The main limitation then becomes any insufficiencies in the specified prompt instructions or tool definitions for a particular feature. When definitions did not provide the proper constraints around a field's intended clinical meaning or the definition contained ambiguity, the model would occasionally disagree with the expert provided labels. Refinement of prompts and tool schema constraints substantially decreased these errors, underscoring the importance of high-quality, detailed, clinically informative specifications.

Interestingly, several cases where the model produced a false positive result were found to accurately represent the clinical attribute detailed in the EHR upon review and were mislabeled in the gold-standard set. This highlights the inherent difficulty of expert chart review and supports the potential of LLM systems for enhancing generation of such datasets on vast EHR repositories. Further, manual extraction of data is highly time-consuming, which often limits the number of observations included in retrospective research. Manual review of 100 charts to generate reference data required an initial investment of approximately 30 minutes per patient chart. Subsequent quality-control review of 10% of the charts by an independent reviewer resulted in an additional 30-minute investment per chart, representing approximately 55 hours of manual effort. Our system can complete the same analysis and quality-control review in less than one-tenth of the time while producing consistent and reproducible outputs.

To increase sample size without laborious manual chart review, many researchers rely on billing codes to identify diagnoses. However, numerous studies have demonstrated that billing codes have limited specificity and sensitivity for complex conditions. Our data set confirms this limitation. Despite selecting patients with international classification of disease codes consistent with traumatic SCI, only 51% of charts manually reviewed reflected true traumatic SCI. These findings underscore the need for tools that achieve both accuracy and efficiency in cohort identification.

As health systems continue their transition to fully electronic medical records, the volume of healthcare data available for research will expand exponentially. Without systems capable of systematically codifying and structuring this data, it will remain inaccessible for clinical research. AI-enabled extraction tools allow for the analysis of large volumes of nuanced clinical information. This data can be utilized to improve predictive modeling, advance our understanding of patient trajectories, and support evidence-based clinical decision making. Development of a secure web-based tool that delivers this level of nuanced health data to clinicians and researchers represents a critical step towards realizing these goals.

**Acknowledgement**
This research was supported in part by the National Institutes of Health under award numbers UL1TR001998 and T32NS077889. The content is solely the responsibility of the authors and does not necessarily represent the official views of the NIH.